\documentclass{article}
\usepackage{spconf,amsmath,graphicx}
\usepackage{booktabs}
\usepackage{multirow}
\usepackage{amssymb}
\usepackage{pifont}
\usepackage{hyperref}
\usepackage{microtype}
\usepackage{marvosym}

\usepackage{enumitem}
\setlist{nosep, leftmargin=14pt}

\usepackage{mwe} % to get dummy images

\title{SurgEye: Instruction Tuning-Free Perception-Reasoning Framework for Surgical Scene Understanding}
\name{
    \begin{tabular}{@{}c@{}}
    Qing Xu$^{1,2*}$, Xiechen Shen$^{1,*}$, Xianjian He$^{1,}$\textsuperscript{\Letter}, Guqiao Liang$^{1,*}$,  Zixiao Chai$^{1,*}$, Jiale Qin$^{1,\dag}$, Yifan Chen$^{1,\dag}$ \\
    Siyi Li$^{1,2,\dag}$, Tian Yuan$^{1,2}$, Yiwei Li$^{1}$, Ziheng Meng$^{1,2}$, Zhen Chen$^{3,}$\textsuperscript{\Letter}\thanks{$^*$Joint first author. $^\dag$ Joint second author. \textsuperscript{\Letter}Corresponding author.}
    \end{tabular}
}
\address{$^1$School of Computer Science, University of Nottingham Ningbo China, $^2$University of Nottingham\\ $^3$ Department of Data Science and Artificial Intelligence, The Hong Kong Polytechnic University}
\begin{document}
%\ninept
%
\maketitle
\begin{abstract}
Surgical scene understanding is pivotal for computer-assisted intervention, yet demands compositional reasoning across instruments, anatomy, and workflow. Existing surgical vision-language models rely on instruction tuning with large-scale question-answer pairs, entailing laborious data curation and costly retraining. To bridge this gap, we propose SurgEye, a perception-reasoning framework that grounds reasoning in a vision foundation model without instruction tuning. Specifically, we devise Routed Multi-Task Perception (RMP) that routes the vision foundation model through task-specific pathways learned from public perception labels, producing spatial and temporal evidence. We further introduce Agentic Compositional Reasoning (ACR) that employs a frozen large language model to coordinate RMP specialists and temporal memory as callable tools, composing evidence across tasks and time into grounded answers. Moreover, we construct SurgEye-Bench spanning eight surgical procedures, pairing each question with perception labels for verifiable evaluation. Extensive experiments demonstrate that SurgEye outperforms state-of-the-arts in surgical reasoning. The source code will be available.
\end{abstract}

\section{Introduction}
\label{sec:intro}

Surgical scene understanding plays a crucial role in computer-assisted intervention \cite{ayobi2025pixel}. Beyond recognizing phases, events, instruments, anatomy, and their interactions \cite{lavanchy2024challenges,nasirihaghighi2025gynsurg,ross2021comparative,hong2020cholecseg8k,nwoye2022rendezvous}, intraoperative decision-making requires reasoning over these observations, \textit{e.g.}, whether bleeding occurs near critical anatomy in the current phase. Such reasoning must compose fine-grained spatial evidence with long-range procedural context, while remaining faithful to the observed scene. These demands have given rise to the challenging task of surgical reasoning, which seeks to derive reliable conclusions from the perceived scene.

The classical surgical understanding methods \cite{nwoye2022rendezvous,lavanchy2024challenges,luo2026rasp} have been designed for specific tasks, resulting in isolated solutions for instrument segmentation, anatomy parsing, and workflow recognition. Meanwhile, vision foundation models (VFMs) self-supervised on large-scale surgical data provide transferable representations \cite{batic2024endovit,JASPERS2026103873,de2026towards}, efficiently adaptable to individual tasks via adapters \cite{hu2022lora}. Despite precise task-specific predictions, these models are constrained to fixed outputs and thus cannot support compositional surgical reasoning.

Recent studies on Vision-Language Models (VLMs) have enabled surgical reasoning in natural language, typically through instruction tuning on large-scale question-answer pairs \cite{sellergren2025medgemma,wang2025endochat,zeng2025surgvlm}. However, constructing such data requires laborious curation, and every extension to new tasks or procedures entails costly retraining of billion-scale models. In fact, VFMs already deliver precise surgical perception from existing labels, offering a natural alternative to relearning such perception through instruction tuning. This motivates us to treat the VFM as the eyes of a frozen large language model (LLM), grounding its reasoning in VFM-based perception.

\begin{figure*}[!t]
  \centering
  \begin{minipage}[b]{0.33\linewidth}
    \centering
    \includegraphics[width=\linewidth]{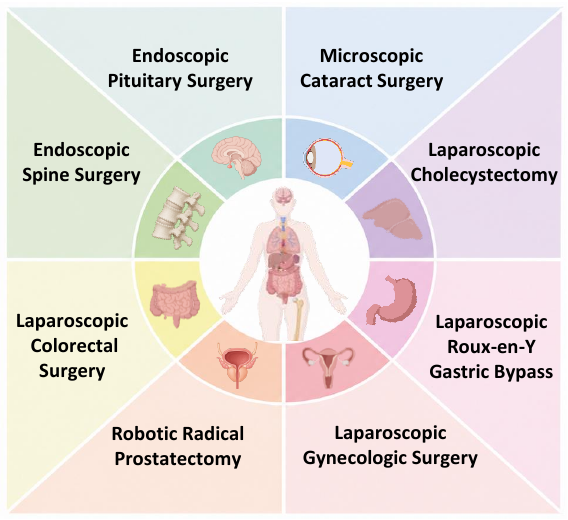}\\[2pt]
    {\small (a) \textbf{Scene}: eight surgical procedures}
  \end{minipage}\hfill
  \begin{minipage}[b]{0.28\linewidth}
    \centering
    \includegraphics[width=\linewidth]{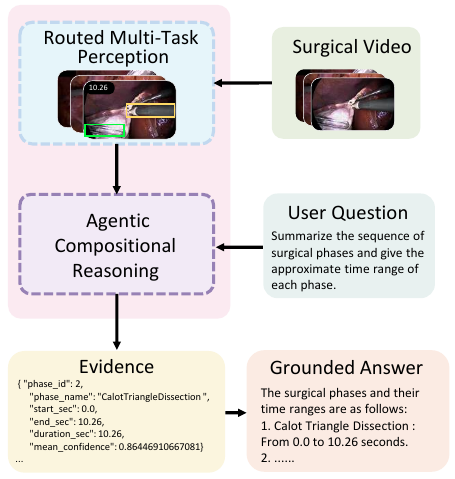}\\[2pt]
    {\small (b) \textbf{Model}: SurgEye}
  \end{minipage}\hfill
  \begin{minipage}[b]{0.375\linewidth}
    \centering
    \includegraphics[width=\linewidth]{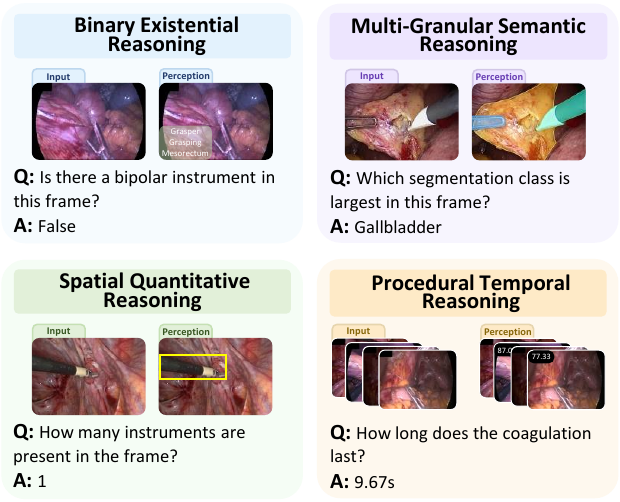}\\[2pt]
    {\small (c) \textbf{Benchmark}: SurgEye-Bench}
  \end{minipage}
  \caption{Illustration of SurgEye for surgical reasoning without instruction tuning. (a) Diverse surgical scenes spanning eight procedures. (b) SurgEye employs a VFM as the eyes of a frozen LLM, grounding answers in structured perception evidence. (c) SurgEye-Bench aligns each question-answer pair with its perception labels for verifiable evaluation across four dimensions.}
  \label{fig:intro}
\end{figure*}

To bridge this gap, we propose SurgEye, a perception-reasoning framework that grounds surgical reasoning in a VFM without instruction tuning, as illustrated in Fig.~\ref{fig:intro}. Specifically, we devise Routed Multi-Task Perception (RMP) that routes the VFM through task-specific pathways learned from public perception labels, producing structured spatial and temporal evidence. We further introduce Agentic Compositional Reasoning (ACR) that employs a frozen LLM to coordinate RMP specialists and structured temporal memory as callable tools, composing cross-task and cross-temporal evidence into grounded answers. This design avoids training-time question-answer curation and keeps every answer verifiable against perception evidence. Moreover, we construct SurgEye-Bench spanning eight surgical procedures, deriving each question from perception labels across four evaluation dimensions.

\section{Methodology}
\label{sec:method}

\begin{figure*}[!t]
  \centering
  \includegraphics[width=1\linewidth]{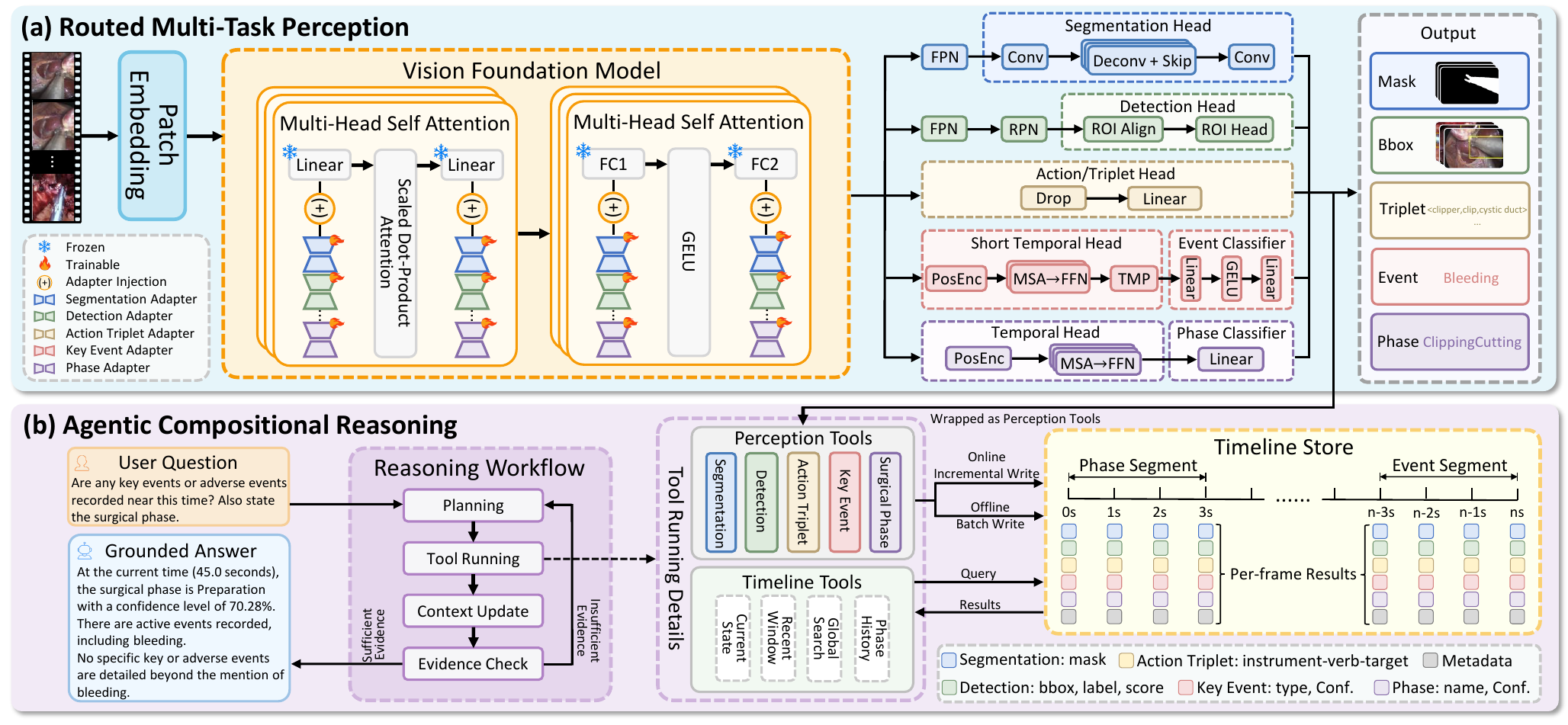}
\caption{The overview of our SurgEye framework for surgical reasoning, consisting of (a) RMP and (b) ACR. SurgEye coordinates VFM-based perception specialists as callable tools for grounded LLM reasoning without instruction tuning.}
  \label{fig:method}
\end{figure*}

We present the SurgEye framework in Fig.~\ref{fig:method}, and achieve surgical reasoning without instruction tuning. To accomplish this, we design the RMP to produce structured surgical evidence and the ACR to coordinate RMP specialists as callable tools for a frozen LLM, with SurgEye-Bench constructed for verifiable evaluation, enabling SurgEye to ground surgical reasoning in reliable VFM-based perception evidence instead of large-scale curated question-answer pairs.

\subsection{Routed Multi-Task Perception}
The recent VFM \cite{simeoni2025dinov3} delivers powerful general-purpose visual representations, yet adapting it to heterogeneous surgical tasks typically requires an independently fine-tuned backbone for each task, which duplicates parameters and isolates perception capabilities. To address this, we devise RMP that routes the shared VFM through task-specific adaptation pathways learned from public perception labels, producing structured surgical evidence across spatial and temporal granularities.

Specifically, given a surgical image or video clip $x$, we insert a task-specific LoRA module with parameters $\theta_k$ into the frozen VFM $\Phi_\theta$ and attach a dedicated decoder $D_k$ for the $k$-th task, as follows:
\begin{equation}
Y(x) = \{\hat{y}_k\}_{k=1}^{N}, \quad \hat{y}_k = D_k\big(\Phi_{\theta,\theta_k}(x)\big),
\label{eq:rmp}
\end{equation}
where $\theta$ denotes the vision encoder parameters shared across all $N$ tasks, and $\hat{y}_k$ denotes the structured prediction of the $k$-th task. Since only $\theta_k$ and $D_k$ are optimized, the same input can be routed through any requested pathway without duplicating the backbone. We instantiate RMP with five surgical perception tasks spanning different spatial and temporal granularities. At the spatial level, the segmentation decoder delineates instruments and anatomical structures at the pixel level, while the detection decoder localizes instruments and associates them across frames for tracking. Beyond individual objects, the action triplet decoder characterizes instrument--verb--target interactions. At the temporal level, the key event decoder aggregates short-range context to identify critical intraoperative events, whereas the phase decoder models long-range dependencies to recognize procedural phases. In this way, RMP provides structured predictions that serve either as task-specific outputs or as grounded visual evidence for ACR.

\subsection{Agentic Compositional Reasoning}
Existing medical and surgical VLMs \cite{sellergren2025medgemma,wang2025endochat} infer surgical states within a single instruction-tuned model, whereas surgical reasoning often requires evidence spanning multiple tasks and time points. To overcome this bottleneck, we introduce ACR that employs a frozen LLM as an agent to coordinate RMP specialists and a structured temporal memory as callable tools, composing cross-task and cross-temporal evidence into grounded answers. Different from VLM-based approaches, our agent never processes raw pixels but reasons entirely over structured perception evidence.

Specifically, given a user question and the input surgical image or video, the agent consults a predefined skill bank that specifies the function, inputs, and outputs of each perception and timeline tool, and accordingly invokes one or multiple RMP specialists on the input, where a compositional question triggers multiple tools whose observations span different tasks. For a surgical video sampled at timestamps $\{t_i\}_{i=1}^{L}$, ACR further organizes the RMP predictions into a structured temporal memory $M$ indexed by time and task, which supports flexible retrieval as follows:
\begin{equation}
\begin{aligned}
M(t_i, k) &= D_k\big(\Phi_{\theta,\theta_k}(x_{t_i})\big), \\
M[W, S_q] &= \big\{M(t_i, k) \mid t_i \in W,\, k \in S_q\big\},
\end{aligned}
\label{eq:memory}
\end{equation}
where $x_{t_i}$ denotes the input frame or clip at $t_i$, $W$ denotes a temporal window, and $S_q$ denotes the subset of tasks required by the question. Timeline tools instantiate $W$ as the current timestamp, a recent window, or the entire procedure to retrieve the current state, recent context, global events, and phase history, so that $M$ is constructed once and reused across subsequent questions without re-executing RMP. The agent then iteratively updates its context with the retrieved observations and checks whether the collected evidence is sufficient, invoking additional tools when necessary. The final answer is generated solely from the collected observations, distinguishing unavailable evidence from confirmed negative findings. In this way, ACR grounds the reasoning of the frozen LLM in VFM-based perception evidence, enabling zero-shot surgical reasoning without any instruction tuning.

\begin{table}[!t]
\centering
\small
\setlength\tabcolsep{4pt}
\caption{Comparison with state-of-the-art methods on surgical reasoning of SurgEye-Bench (Accuracy, \%). $^{\star}$ denotes supervised fine-tuning on the training split of SurgEye-Bench.}
\resizebox{\columnwidth}{!}{\begin{tabular}{l|c|cccc}
\hline
Methods & \#Params & MSR & BER & SQR & PTR \\
\hline
Qwen3-VL \cite{bai2025qwen3vl} & 30B & 33.38 & 50.68 & 33.06 & 20.55 \\
Qwen3.8 \cite{qwen2026qwen38} & 27B & 28.82 & 48.43 & 40.30 & 17.88 \\
Gemma 4 \cite{team2026gemma} & 26B & 51.58 & 55.77 & 54.78 & 19.46 \\
MedGemma \cite{sellergren2025medgemma} & 27B & 34.42 & 54.48 & 29.99 & 20.47 \\
EndoChat \cite{wang2025endochat} & 13B & 1.62 & 52.36 & 20.17 & 7.15 \\
\hline
Qwen3-VL$^{\star}$ \cite{bai2025qwen3vl} & 30B & 57.19 & 65.53 & 49.92 & 46.07 \\
\hline
SurgEye & 0.3B+30B & \textbf{76.05} & \textbf{76.78} & \textbf{56.21} & \textbf{78.11} \\
\hline
\end{tabular}}
\label{tab:reasoning}
\end{table}

\subsection{Perception-Reasoning Paired Benchmark}
To jointly evaluate surgical perception and reasoning, we construct SurgEye-Bench, where each question is derived from and paired with the ground-truth perception labels of the same sample. The annotations are collected from nine public datasets \cite{hong2020cholecseg8k,ross2021comparative,xu2025spine,nwoye2022rendezvous,ayobi2025pixel,nasirihaghighi2025gynsurg,lavanchy2024challenges,ghamsarian2024cataract,das2025pitvis} spanning eight surgical procedures, covering segmentation, detection, action triplet, key event, and phase recognition. Specifically, these annotations are unified into the image or video clip, task labels, and derived attributes (\textit{e.g.}, area ratios, instrument counts, and event intervals), from which question-answer pairs are automatically generated. This pairing makes every answer verifiable, allowing reasoning errors to be traced back to perception evidence.

\begin{table}[!t]
\centering
\small
\setlength\tabcolsep{9pt}
\caption{Ablation study of SurgEye on SurgEye-Bench. }
\resizebox{\columnwidth}{!}{\begin{tabular}{c|cc|cccc}
\hline
\# & RMP & ACR & MSR & BER & SQR & PTR \\
\hline
1 & & & 25.72 & 50.15 & 8.08 & 0.00 \\
2 & \checkmark & & 32.71 & 63.96 & 18.85 & 2.89 \\
% 2 & \checkmark & & 24.72 & 54.35 & 7.00 & 3.75 \\
% 3 & & \checkmark & 64.62 & 68.77 & 44.41 & 48.89 \\
3 & \checkmark & \checkmark & 76.05 & 76.78 & 56.21 & 78.11 \\
\hline
\end{tabular}}
\label{tab:ablation}
\end{table}

On this basis, we organize the question-answer pairs into four evaluation dimensions. Multi-Granular Semantic Reasoning (MSR) and Binary Existential Reasoning (BER) span all five tasks, evaluating the identification of surgical entities, actions, events, and phases, and the presence of a specified target, respectively. Spatial Quantitative Reasoning (SQR) evaluates frame-level area ratios, instrument counts and locations, and action counts, whereas Procedural Temporal Reasoning (PTR) evaluates event durations and phase statistics. Overall, SurgEye-Bench comprises 63K QA samples over 452K surgical frames, including 17.1K, 20.0K, 12.6K, and 13.7K samples for MSR, BER, SQR, and PTR, respectively.

\begin{figure}[!t]
  \centering
  \includegraphics[width=1\linewidth]{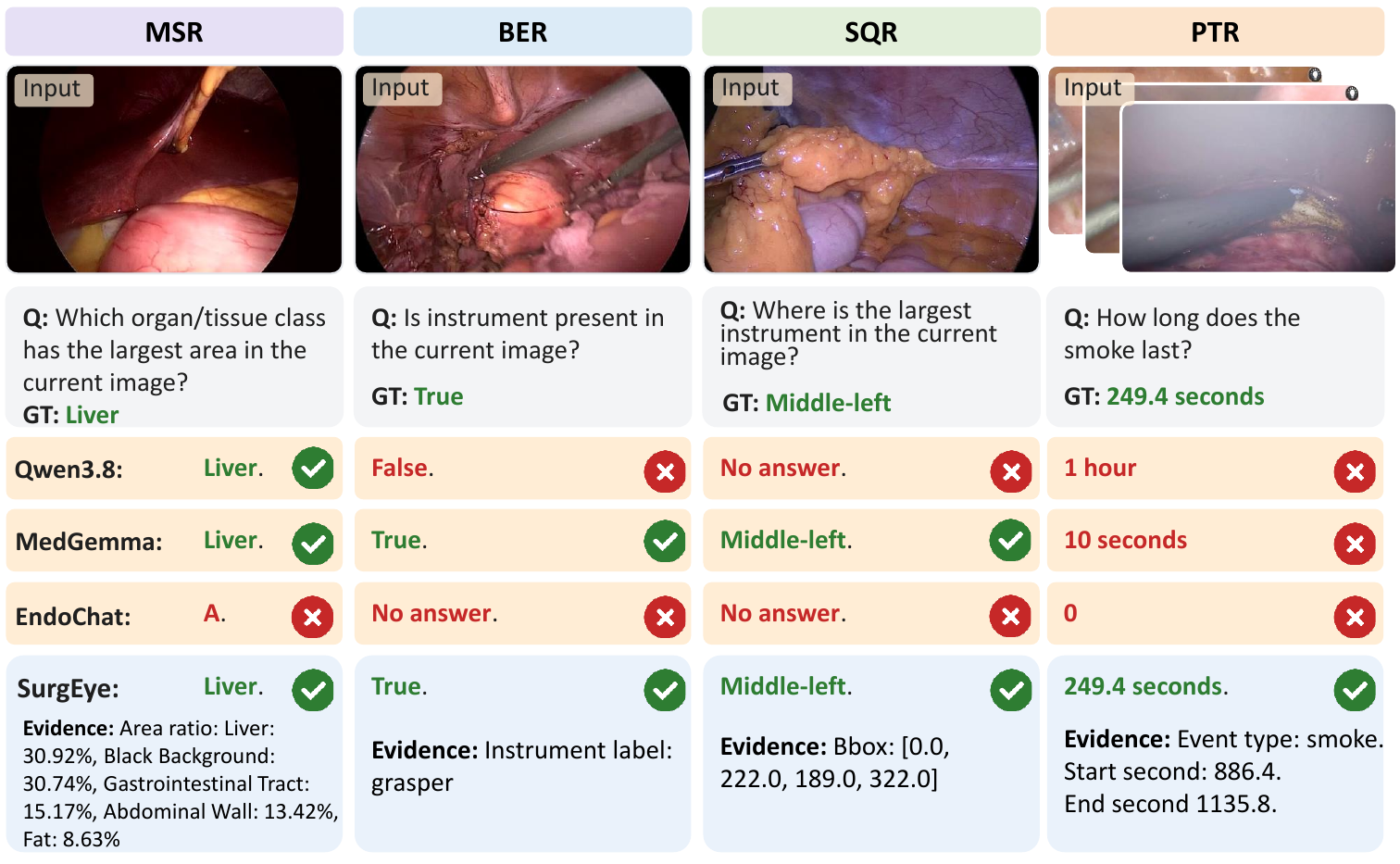}
  \caption{Qualitative comparisons across four evaluation dimensions of SurgEye-Bench. Our SurgEye exhibits the best results, producing correct answers on all four dimensions, with perception evidence that makes each answer verifiable.}
  \label{fig:qualitative}
\end{figure}

\begin{table}[!t]
\centering
\small
\setlength\tabcolsep{6pt}
\caption{Comparison with state-of-the-art vision foundation models on surgical perception, averaged over datasets of each task. SEG: segmentation (Dice), DET: detection (AP$_{50}$), ATR: action triplet recognition (AP$_{\rm IVT}$), KER: key event recognition (F1), PHR: phase recognition (F1).}
\resizebox{\columnwidth}{!}{\begin{tabular}{l|ccccc}
\hline
Methods & SEG & DET & ATR & KER & PHR \\
\hline
ViT-L \cite{dosovitskiy2021an} & 79.40 & 69.53 & 32.39 & 92.54 & 71.18 \\
EndoViT \cite{batic2024endovit} & 79.31 & 67.79 & 33.99 & \textbf{93.91} & 58.65 \\
GSViT \cite{schmidgall2024general} & 60.19 & 44.25 & 25.03 & 82.35 & 21.11 \\
SurgeNet-XL \cite{JASPERS2026103873} & 79.51 & 74.61 & 37.22 & 92.82 & 66.76 \\
SurgeNetDINO-L \cite{de2026towards} & 81.23 & 79.74 & \textbf{39.23} & 93.56 & 67.69 \\
\hline
SurgEye & \textbf{82.59} & \textbf{82.66} & 35.76 & 91.02 & \textbf{72.40} \\
\hline
\end{tabular}}
\label{tab:perception}
\end{table}

\section{Experiments}
\label{sec:format}

\subsection{Experimental Setup}
We perform all experiments on four NVIDIA L20 GPUs using PyTorch. We adopt DINOv3 ViT-L \cite{simeoni2025dinov3} as the frozen VFM in RMP and Qwen3-30B-A3B \cite{yang2025qwen3} as the frozen LLM agent in ACR. All images and video frames are resized to $384 \times 384$, and RMP is optimized by AdamW with an initial learning rate of $1\times10^{-4}$ and cosine annealing with 5\% warm-up. We split all datasets at the video level into the training, validation, and test sets as 7:1:2. For perception, we adopt Dice, AP$_{50}$, AP$_{\rm IVT}$ \cite{nwoye2022rendezvous}, and F1 score for segmentation, detection, action triplet recognition, and key event and phase recognition, respectively. For reasoning, we report the accuracy on each evaluation dimension of SurgEye-Bench.

\subsection{Comparison with State-of-the-Art Methods}
To comprehensively evaluate the performance of SurgEye, we first perform comparisons with state-of-the-art VLMs on SurgEye-Bench. As shown in Table~\ref{tab:reasoning}, zero-shot VLMs remain close to chance level on BER, while MedGemma \cite{sellergren2025medgemma} and EndoChat \cite{wang2025endochat} show no consistent advantage over general-purpose VLMs \cite{bai2025qwen3vl,qwen2026qwen38,team2026gemma} despite domain-specific instruction tuning. Although supervised fine-tuning substantially improves Qwen3-VL$^{\star}$, it still lags behind SurgEye with the same 30B LLM scale. In contrast, SurgEye achieves 76.05\%, 76.78\%, 56.21\%, and 78.11\% accuracy on MSR, BER, SQR, and PTR, surpassing Qwen3-VL$^{\star}$ by 18.86\%, 11.25\%, 6.29\%, and 32.04\%, respectively. Moreover, Fig.~\ref{fig:qualitative} shows that SurgEye produces correct answers with verifiable perception evidence across all four dimensions.

\subsection{Ablation Study}
To investigate the effectiveness of the two core modules in SurgEye, we conduct ablation studies on SurgEye-Bench, as illustrated in Table~\ref{tab:ablation}. By removing both RMP and ACR, we construct the baseline ($1^{st}$ row), where the frozen LLM answers questions from text alone without visual input. By introducing RMP ($2^{nd}$ row), whose predictions are directly serialized into the prompt, the average accuracy improves by 8.61\%, yet PTR remains only 2.89\%, as the LLM is distracted by massive task-irrelevant predictions. On this basis, our complete SurgEye ($3^{rd}$ row) further employs ACR to coordinate perception tools and structured temporal memory, bringing an average improvement of 42.19\%, with the most significant gains on PTR (+75.22\%) and SQR (+37.36\%). These results validate that RMP and ACR collectively contribute to the superior performance of SurgEye.

\subsection{Multi-Task Perception Analysis}
To evaluate the perception evidence underlying ACR, we further compare SurgEye with state-of-the-art VFMs on surgical perception in Table~\ref{tab:perception}. SurgEye achieves the best segmentation, detection, and phase recognition performance with 82.59\% Dice, 82.66\% AP$_{50}$, and 72.40\% F1, surpassing SurgeNetDINO-L \cite{de2026towards} by 1.36\% Dice and 2.92\% AP$_{50}$. Meanwhile, SurgEye remains competitive with surgical VFMs on action triplet and key event recognition, while sharing a single frozen VFM across all five tasks without duplicating backbones. These results confirm that our SurgEye framework provides reliable perception evidence for ACR.

\section{Conclusion}
In this work, we present SurgEye, which employs a VFM as the eyes of a frozen LLM to achieve surgical reasoning without instruction tuning. We devise RMP to produce structured surgical evidence, and ACR to coordinate RMP specialists and structured temporal memory as callable tools for grounded reasoning. Additionally, we construct SurgEye-Bench, pairing perception labels with reasoning questions. Extensive experiments demonstrate that our SurgEye outperforms state-of-the-art methods in surgical reasoning and perception.

% \section{Acknowledgements}
% This work is partially supported by the Yongjiang Technology Innovation Project (2022A-097-G), Zhejiang Department of Transportation General Research and Development Project (2024039), National Natural Science Foundation of China grant (62476037), and Ningbo Science and Technology Bureau under NBNSF General Programme with (2025J114). 

% References should be produced using the bibtex program from suitable
% BiBTeX files (here: strings, refs, manuals). The IEEEbib.bst bibliography
% style file from IEEE produces unsorted bibliography list.
% ------------------------------------------------------------------------- 
{\small
\bibliographystyle{IEEEbib}
\bibliography{refs}
}

\end{document}